\pdfoutput=1

\documentclass[11pt]{article}

\usepackage[]{acl}

\usepackage{times}
\usepackage{latexsym}

\usepackage[T1]{fontenc}

\usepackage[utf8]{inputenc}

\usepackage{microtype}

\usepackage{algorithm}
\usepackage{amsmath}
\usepackage{arydshln}
\usepackage{algpseudocode}
\usepackage{bm}
\usepackage{graphicx}
\usepackage{multirow}
\usepackage{booktabs}

\usepackage{subcaption}
\usepackage{graphicx}
\usepackage{enumitem}

\title{A Pilot Study on Dialogue-Level Dependency Parsing for Chinese}

\author{Gongyao Jiang\textsuperscript{1}, Shuang Liu\textsuperscript{2}, Meishan Zhang\textsuperscript{3}\thanks{ \ \ Corresponding author} , Min Zhang\textsuperscript{3} \\
  \textsuperscript{1}School of New Media and Communication, Tianjin University, China\\
  \textsuperscript{2}College of Intelligence and Computing, Tianjin University, China \\
  \textsuperscript{3}Institute of Computing and Intelligence, Harbin Institute of Technology (Shenzhen), China \\
  \texttt{jianggongyao@gmail.com, shuang.liu@tju.edu.cn}\\ 
  \texttt{mason.zms@gmail.com,  zhangmin2021@hit.edu.cn}\\
  }

\begin{document}
\maketitle
\begin{abstract}
Dialogue-level dependency parsing has received insufficient attention, especially for Chinese. To this end, we draw on ideas from syntactic dependency and rhetorical structure theory (RST), developing a high-quality human-annotated corpus, which contains 850 dialogues and 199,803 dependencies. Considering that such tasks suffer from high annotation costs, we investigate zero-shot and few-shot scenarios. Based on an existing syntactic treebank, we adopt a signal-based method to transform seen syntactic dependencies into unseen ones between elementary discourse units (EDUs), where the signals are detected by masked language modeling. Besides, we apply single-view and multi-view data selection to access reliable pseudo-labeled instances. Experimental results show the effectiveness of these baselines. Moreover, we discuss several crucial points about our dataset and approach.
\end{abstract}

\section{Introduction}
As a fundamental topic in the natural language processing (NLP) community, dependency parsing has drawn a great deal of research interest for decades \citep{marcus-etal-1994-penn, mcdonald-etal-2013-universal, zhang-2021-dependency}.
The goal of dependency parsing is to find the head for each word and the corresponding relation \citep{kubler-2009-dependency}.
Most of previous works have focused on the sentence level, while the dialogue-level dependency parsing still stands with the paucity of investigation.

Prior studies build dialogue-level discourse parsing datasets \citep{asher-etal-2016-discourse, li-etal-2020-molweni} with reference to the text-level discourse dependency \citep{li-etal-2014-text}.
The discourse structures in these data are constructed by elementary discourse units (EDUs) and the relationships between them, without regard to the inner structure of EDUs.
It is of interest to incorporate both inner-EDU and inter-EDU dependencies throughout a dialogue to construct a word-wise dependency tree, which is in line with sentence-level dependencies and further able to express hierarchical structures.
Hence, we form the dialogue-level dependency, by adapting commonly-used syntactic dependency \citep{jiang-etal-2018-supervised} and rhetorical structure theory (RST) dependency \citep{carlson-etal-2001-building, li-etal-2014-text} into inner-EDU and inter-EDU dependencies.
\begin{figure}[t]
    \centering
    \includegraphics[width=0.49\textwidth]{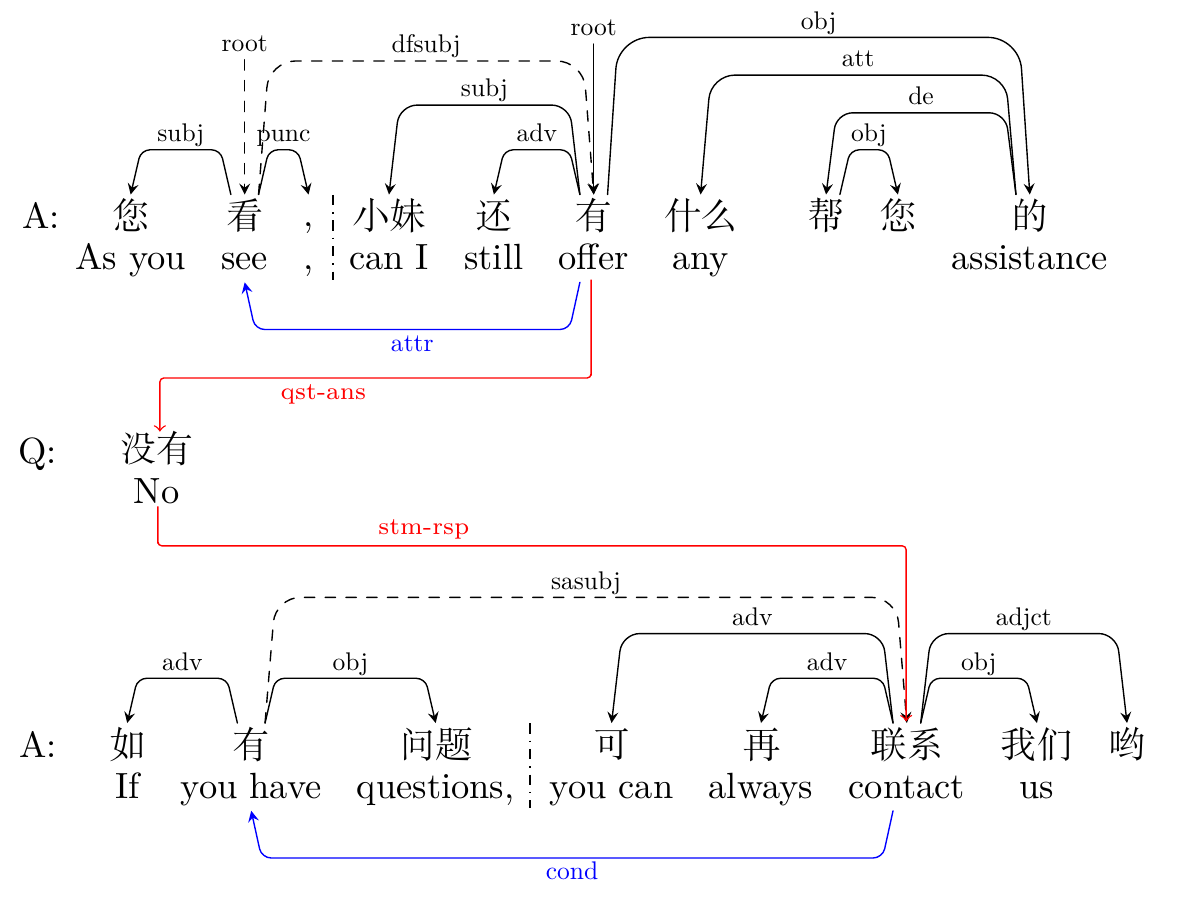}
    \caption{A fragment example of dialogue-level dependencies. Vertical dashed lines are separation boundaries of EDUs. Above words are the inner-EDU dependency arcs, while the arcs below words or cross utterances represent the inter-EDU dependencies.}
    \label{fig:example}
\end{figure}

The scarcity of research on dialogue-level dependency parsing might be caused by the prohibitive cost of annotation.
Thus, we focus on low-resource settings, aiming to construct a sufficient test set to support reliable evaluation and label a small training set.
We craft an annotation guideline and develop a platform for human labeling.
After that, we perform manual annotation and obtain 50 training instances and 800 test samples.
Figure \ref{fig:example} illustrates a fragment in a labeled dialogue, which contains three utterances with dependency links.

Learning a parser with scant to no supervised signals can be challenging.
Fortunately, it is straightforward to use an existing syntactic treebank to attain inner-EDU dependencies.
Meanwhile, we find some overlap between syntactic dependency and inter-EDU dependency.
As shown in Figure \ref{fig:example}, ``offer'' is dependent on ``see'' with the ``dfsubj'' (different subject) relationship in syntactic dependency, which matches the ``attr'' (attribution) of inter-EDU dependency.
Furthermore, inter-EDU arcs between utterances are often emitted from the root node above to the root below.
Hence, we can naturally obtain inter-EDU arcs from partial syntactic dependencies.

We find that certain words can reflect the discourse role of EDUs, helping to assign labels to inter-EDU arcs.
For instance, ``see'' and ``if'' reflect the ``attribution'' and ``condition'' signals respectively, illustrated in Figure \ref{fig:example}.
Thus, we propose a method to discover and leverage these signals for inter-EDU label assignment.
Inspired by prompt-based learning \citep{liu-etal-2021-prompt}, we adopt masked language modeling (MLM) to recover signal words, the set of which is carefully defined in advance.
Predicted words are then mapped to corresponding signals.
Based on these signals, a handful of simple well-defined rules can infer credible dependencies.

The gap between the syntactic treebank and dialogue-level dependencies goes beyond labels; differences in text style expose a trained parser to the out-of-distribution (OOD) effect \citep{li-etal-2019-semi-supervised-domain}.
To alleviate that, we utilize unlabeled dialogue data to bridge the cross-domain gap.
We borrow ideas from self-training \citep{scudder-1965-probability} and co-training \citep{blum-1998-combining}, applying single-view and multi-view data filtering augmentation.
Given pseudo-labeled samples predicted by parsers, we calculate their confidence scores and set a threshold.
Instances above the threshold are provided as additional training samples.

We carry out experiments in zero-shot and few-shot scenarios.
The empirical results suggest that our signal-based baseline can achieve reasonable performance in the zero-shot condition and provide significant improvement in the few-shot setting.
Incorporating filtered pseudo-labeled data further advances performance.
In addition, we present several discussion topics, which can affirm directive intuitions and crucial practices in this work. 

Our contributions are mainly in three respects:
\begin{itemize}[noitemsep, topsep=0pt]
    \item We build the first Chinese dialogue-level dependency treebank.
    \item We propose signal-based dependency transformation and pseudo-labeled data filtering approaches as parsing baselines.
    \item The proposed methods achieve reasonable parsing performance in low-resource settings.
\end{itemize}
All our datasets and codes will be publicly available at \href{https://github.com/Zzoay/DialogDep}{github.com/Zzoay/DialogDep} for research purpose. 
\section{Dataset Construction}
To facilitate our research on dialogue-level dependency parsing,
here we construct a high-quality corpus manually, named the Chinese dialogue-level dependency treebank (CDDT).
The treebank borrows the sentence-level dependency \citep{jiang-etal-2018-supervised} and the discourse-level dependency \citep{li-etal-2014-text}\footnote{Appendix \ref{sec:rst} provides a brief introduction to this discourse-level parsing schema, i.e., RST.}, extending both to the dialogue texts.
Below, we present the annotation details.

\subsection{Annotation Preparation}
\noindent
\textbf{Data Collection.}
We use a publicly available dialogue dataset \citep{chen-etal-2020-jddc} as our raw corpus, which is collected from real-world customer service, containing multiple turns of utterance.
We access this data from an available source.\footnote{
As the public link to this data is inaccessible, we utilize another version made public by CSDS \citep{lin-etal-2021-csds}, a dialogue summary dataset, based on the same dialogue data.}
We use 800 data from the full test set as the raw corpus of our test set and randomly sample 50 data from the 9101 training data as our raw corpus for few-shot learning.
The data providers have segmented words and anonymized private information.
We further manually clean the data, removing noise text and performing fine-grained word segmentation.

\noindent
\textbf{Guideline.}
A well-defined dialogue-level dependency tree can be decomposed into two parts, i.e., inner-EDU dependencies and inter-EDU dependencies, respectively.
Following \citet{carlson-etal-2001-building}, we treat EDUs as non-overlapping spans of text and use lexical and syntactic clues to help determine boundaries.
To adapt to dialogue parsing and linguistic features of Chinese, we simplify the boundary determination, regarding the boundaries as the leftmost and rightmost words of spans, whose words are covered by a complete syntactic subtree.
The root nodes of each subtree are often predicates that reflect single semantic events, illustrated in Figure \ref{fig:example}.

For the inner-EDU dependency annotation, we follow the guideline proposed by \citet{jiang-etal-2018-supervised}, which includes 21 classes of syntactic dependency.
For the inter-EDU one, we examine and adapt the EDU-wise relations \citep{carlson-etal-2001-building} to word-wise dependencies, including 15 coarse-grained relations and 4 fine-grained relations divided from ``topic-comment'' to accommodate connections between utterances.
In annotation, inner-EDU dependencies should be annotated first as the underlying basis, while inter-EDU dependencies annotation should take full account of the semantic relations between EDUs.
Appendix \ref{sec:dep} shows all the dependencies and reveals more details about our dialogue-level dependencies.

\subsection{Human Annotation}
\noindent
\textbf{Platform.}
The dialogue-level dependency annotation consists of inner-utterance and inter-utterance parts, where the former is in line with the common sentence-level annotation, and the latter needs to connect from above to below.
To accommodate this form of human labeling, we develop an online annotation platform.
Annotators can easily label the inner-utterance and inter-utterance dependencies on this platform by linking tags between words.

\noindent
\textbf{Annotation Process.}
We employ two postgraduate students as annotators, both of whom are native Chinese speakers.
Each annotator is intensively trained to familiarise the guideline.
To guarantee the labeling quality, we apply a double annotation and expert review process.
Each annotator is assigned to annotate the same raw dialogue.
The annotation is accepted if the two submissions are the same.
Otherwise, an experienced expert with a linguistic background will compare the two submissions and decide on one answer.
Moreover, the expert checks all labeling for quality assurance.

\noindent
\textbf{Statistics.}
\begin{table}[t]
\centering
\begin{tabular}{lrr}
\hline
\textbf{Statistic} &  \textbf{Train} & \textbf{Test}\\
\hline
\# dialogue & 50 & 800 \\
avg.\# turns & 23 & 25 \\
avg.\# words & 194 & 212 \\
\# inner & 9129 & 159803 \\
\# inter & 1671 & 29200 \\
\hline
\end{tabular}
\caption{Some statistical information about CDDT. ``\#'' and ``avg.\#'' represent ``the number of'' and ``the average number of'' respectively.}
\label{tab:statistics}
\end{table}
In the end, we arrive at 800 number of annotated dialogues for testing and 50 for few-shot training. Table \ref{tab:statistics} shows some statistical information. 
It can be seen that the number of dialogue rounds is large, leading to an expensive labeling process.
Also, inter-EDU relations are much more scarce compared to inner-EDU ones, as inter-EDU dependencies are higher-level relations that reflect the connections between EDUs.
Appendix \ref{sec:dep} shows the quantities of each category.



\section{Methodology}
Dialogue-level dependency parsing involves two folds, namely inner-EDU dependency parsing and inter-EDU dependency parsing.
We leverage an existing syntactic treebank $\mathcal{S}$ \citep{li-etal-2019-semi-supervised-domain} to learn a parser, which can analyze inner-EDU dependencies well.
Meanwhile, it can be observed that syntactic dependencies have the potential to be converted into inter-EDU ones (Figure \ref{fig:example}).
Thus, we propose a signal-based method to perform dependency transformation for parser training and inference.
Moreover, we apply pseudo-labeled data filtering to leverage the large-scale unlabeled dialogue data $\mathcal{D}$.

\subsection{Signal-based Dependency Transformation}
Our dependency parsing model contains an encoder layer that uses a pretrained language model (PLM) and a biaffine attention decoder as detailed in \citet{dozat-etal-2017-biaffine}.
Given a sentence (in $\mathcal{S}$) or utterance (in dialogues) $\boldsymbol{x}=[w_1, w_2, \ldots, w_n]$, the parser can output the arc and label predictions:
\begin{equation}
\bar{\boldsymbol{y}}^{(arc)}, \bar{\boldsymbol{y}}^{(label)} = \textrm{Parser}\left(\boldsymbol{x}\right)
\end{equation}

In summary, the dependency transformation can be applied in two ways. 
The first is replacing partial labels of predicted syntactic dependencies with inter-EDU ones, similar to post-processing.
We denote it to $\textit{PostTran}$.
The second is to transform the syntactic dependency dataset $\mathcal{S}$ into the one with extended inter-EDU dependencies for parser training, denoted to $\textit{PreTran}$.
For clarity, we denote the transformed treebank to $\mathcal{S}_\tau$.
We refer the parser trained on the syntactic treebank $\mathcal{S}$ to $\textrm{Parser-S}$, and the parser based on $\mathcal{S}_\tau$ to $\textrm{Parser-T}$.

\noindent
\textbf{EDU Segmentation.}
When $\textit{PreTran}$, a sentence in $\mathcal{S}$ is firstly separated into several EDUs.
We exploit a simple but effective way, treating punctuation such as commas, full stops, and question marks as boundaries between EDUs.\footnote{In our preliminary experiments, this method achieves a 68.82\% F1 score of segmentation in those utterances with multiple EDUs, while the single EDU predictions' F1 can be achieved to 90.10\%. We show all the segmentation operators in the publicly available code.}
Meanwhile, some specific syntactic labels (e.g. sasubj, dfsubj) that span a range can also be considered as implicit EDU segmenting signals (Figure \ref{fig:example}).
When $\textit{PostTran}$, an utterance in dialogue is segmented in the same way.
Additionally, there are clear boundaries between utterances, which can be used to separate EDUs.

Given the segmented EDUs, the next issue is where the inter-EDU arcs fall and in what direction.
We find that inside utterances, inter-EDU dependencies are often overlapped with certain syntactic dependencies (Figure \ref{fig:example}).
We name the labels of these syntactic labels as ``transforming labels'' and pre-define a set of them $L = \{root, sasubj, dfsubj\}$.
If $\boldsymbol{y}^{(label)}_i \in L$, we can directly retain (sometimes reverse) the arcs and convert the labels to inter-EDU ones.
Besides, there is no predicted relationship between utterances.
We find that most inter-utterance arcs are emitted from the root node above to the root node below. 
Thus, we link the predicted roots from above to below as the inter-EDU arcs between utterances.

\noindent
\textbf{MLM-based Signal Detection.}
Given those inter-EDU arcs, we propose a signal-based transform approach to assign inter-EDU labels to them.
First, we introduce the signal detection method, which is based on a masked language modeling (MLM) paradigm.
We find that certain words reflect the semantic role of EDUs.
For instance, an EDU that contains ``if'' is commonly the ``condition'' role (Figure \ref{fig:example}).
We can emit an arc ``cond'' from its head EDU to it.
Thus, we pre-define a word-signal dictionary, mapping words to the corresponding inter-EDU relation signals.
A word that reflects the signal is called ``signal word''.

Subsequently, we apply the MLM on the large-scale unlabeled dialogue data $\mathcal{D}$ to learn inter-EDU signals.
During the training stage, the signal word $v$ is randomly dropped in a segmented EDU $\boldsymbol{e}$, and a masked language model is to recover it.
Like prompt-based learning \citep{liu-etal-2021-prompt}, we modify $\boldsymbol{e}$ by a slotted template into a prompt $\boldsymbol{e'}$.
The slot is a placeholder with ``[mask]'' tokens.
Next, a model with PLM and MLP decoder outputs the word distribution in masked positions,\footnote{Since signal words may contain multiple characters in Chinese, the masked place should span several [mask] tokens. 
We average the output probabilities. 
The implemented details are offered in Appendix \ref{sec:imp_mlm}.} as distribution of signal words:
\begin{equation}
P\left(v\middle|\boldsymbol{e}\right) =  \textrm{softmax}\left(\textrm{MLP}\left(\textrm{PLM}\left(\boldsymbol{e'}\right)\right)\right)
\end{equation}
where the $\textrm{MLP}$ is to project the hidden vector to vocabulary space.

At the inference stage, the model outputs the distribution of signal words.
The probabilities of signal words are grouped and averaged by their corresponding signals.
The grouped probabilities form the distribution of inter-EDU signals.
\begin{equation}
P\left(s\middle|\boldsymbol{e}\right) = \textrm{Group\-Mean}\left(P\left(v\middle|\boldsymbol{e}\right)\right)
\end{equation}
The signal $s$ can be obtained by $\textrm{argmax} \ P\left(s|\boldsymbol{e}\right)$.
We expand the predicted signal to the whole $\boldsymbol{e}$.
By executing the above procedure in batch with $\boldsymbol{e}$ in $\boldsymbol{x}$, we end up with the signal sequence, which is denoted in bold $\boldsymbol{s}$.

\noindent
\textbf{Signal-based Transformation.}
Given detected signals $\boldsymbol{s}$, Algorithm \ref{alg:proc} show how partial syntactic labels are transformed into inter-EDU labels.
We pre-set 3 conditions and the corresponding strategies to cover the majority of cases.
1) If the head node of the current word crosses the EDU's boundary, or $\text{the label} \  \boldsymbol{y}^{(label)}_i \in L$ and the connection spans $k$,
we assign the label by the corresponding signal.
2) If $\boldsymbol{s}_i \text{ is ``cond'' or ``attr''}$, we reverse the connections between head and tail nodes.
3) If greetings are the root EDU, then we invert the links, since we are more interested in the core semantics.
The \textit{FindTail} function in Algorithm \ref{alg:proc} performs an iterative procedure. 
It looks for the first node in other EDUs whose head node has an index equal to the current index $i$.
Meanwhile, the labels of arcs between utterances are directly assigned by the signals of their head words.

\begin{algorithm}[t]
\begin{algorithmic}[1]
\Require a sentence (utterance) $\boldsymbol{x}$; arcs $\boldsymbol{y}^{(arc)}$; labels $\boldsymbol{y}^{(label)}$; inter-EDU signals $\boldsymbol{s}$; set of transforming labels $L$; sequence length $n$.
\For {$i=1$ to $n$}
    \If {condition 1} \Comment{certain labels}
    
    \State $\boldsymbol{y}^{(label)}_i \gets \boldsymbol{s}_i$
    
        \If {condition 2}  \Comment{special cases}
        
        \State $t \gets FindTail(\boldsymbol{y}_i^{arc})$
        \State $\boldsymbol{y}^{(arc)}_t \gets \boldsymbol{y}^{(arc)}_i$
        \State $\boldsymbol{y}^{(arc)}_i \gets t$
        
        
        \ElsIf {condition 3}   \Comment{greetings}
        
        \State $t \gets FindTail(\boldsymbol{y}_i^{(arc)})$
        \State $\boldsymbol{y}^{(arc)}_i, \boldsymbol{y}^{(label)}_i \gets t, \text{``elbr''}$
        \State $\boldsymbol{y}^{(arc)}_t, \boldsymbol{y}^{(label)}_t \gets 0, \text{``root''}$
        
        \EndIf
    
    
    \EndIf
    
\EndFor \\
\Return $\boldsymbol{y}^{(arc)}$, $\boldsymbol{y}^{(label)}$
\end{algorithmic}
\caption{Signal-based Transformation}
\label{alg:proc}
\end{algorithm}

\subsection{Unlabeled Data Utilization}
We leverage large-scale dialogue data $\mathcal{D}$ to bridge the distribution gap between the syntactic treebank and our annotated corpus.
It is straightforward to use a well-trained parser to assign pseudo labels to $\mathcal{D}$ for subsequent training.
Nevertheless, this pseudo-labeled data includes a sea of mislabels, which in turn jeopardize performance.
Thus, we apply single-view and multi-view methods to select high-quality additional samples.

\noindent
\textbf{Single-view.} In intuition, predicted samples with higher confidence are of higher quality. 
Given an unlabeled utterance $\boldsymbol{x}$, a well-trained parser ($\textrm{Parser-S}$ or $\textrm{Parser-T}$) can output the probabilities of each arc and label, without the final decoding:
\begin{equation}
    P\left(\boldsymbol{y}^{(arc)}\middle|\boldsymbol{x}\right), P\left(\boldsymbol{y}^{(label)}\middle|\boldsymbol{x}\right) = \textrm{Parser}^* \left(
    \boldsymbol{x}\right)
\end{equation}

Then, we average the highest probabilities in each position, as the confidence of one utterance.
Equation \ref{eq:conf} shows the calculation of arc confidence $c^{(arc)}$.
The label confidence $c^{(label)}$ is computed in the same way.
\begin{equation}
    \label{eq:conf}
    c^{(arc)} = \frac{1}{n} \sum_i^n{\max P\left(\boldsymbol{y}_i^{(arc)}\middle|\boldsymbol{x_i}\right)}
\end{equation}
A pseudo-labeled utterance is reserved when its $c^{(arc)}$ and $c^{(label)}$ both greater than a confidence threshold $\epsilon$.
The filtered samples are incorporated with $\mathcal{S}$ or $\mathcal{S}_\tau$ for training a new parser.

\noindent
\textbf{Multi-view.}
Prior research \citep {blum-1998-combining} and our pilot experiment suggest that multi-view productions of pseudo data outperform single-view ones.
Thus, we exploit $\textrm{Parser-S}$ and $\textrm{Parser-T}$ to conduct multi-view data augmentation.

The confidence computation and filtering methods are the same as the single-view ones above.
The filtered samples labeled by $\textrm{Parser-T}$ and $\textrm{Parser-S}$ are merged together and de-duplicated by their confidence magnitude (i.e. if two samples are the same but have different labels, the one with higher confidence will be retained).
Then, the merged instances are added to $\mathcal{S}$ or $\mathcal{S}_\tau$ for subsequent training.
In this way, parsers trained on the set with such data augmentation can access complementary information.

\section{Experiment}

\subsection{Settings}
\label{sec:setting}

\noindent
\textbf{Evaluation.} We use the labeled attachment score (LAS) and the unlabeled attachment score (UAS) for evaluation.
For a fine-grained analysis, we report the scores of inner-EDU and inter-EDU dependencies.
In the absence of the development set in our low-resource scenarios, we retain the last training checkpoint for evaluation.
To maintain a balance between energy savings and results reliability, in zero-shot settings, we set a random seed of 42 for all experiments and report the testing results.
In few-shot settings, we repeat data sampling on 5 random seeds in 4 few-shots (5, 10, 20, 50) settings.
Then we choose the seeds that obtain median scores for training and final evaluation.
All experiments are carried out on a single GPU of RTX 2080 Ti.

\noindent
\textbf{Hyper-parameters.} Our PLM is a Chinese version of ELECTRA \citep{clark-etal-2020-electra}, implemented by \citet{cui-etal-2020-revisiting}.
We exploit the base scale discriminator\footnote{\href{https://huggingface.co/hfl/chinese-electra-180g-base-discriminator}{huggingface.co/hfl/chinese-electra-180g-base-discriminator}} for fine-tuning.
The hidden size of the subsequent parts of our Parser and MLM is set to 400, and the dropout is 0.2.
We train our models by using the AdamW optimizer \citep{los-2017-adamw}, setting the initial learning rate of the PLM to 2e-5 and of the subsequent modules to 1e-4, with a linear warmup for the first 10\% training steps.
The weight decay is 0.01.
We apply the gradient clipping mechanism by a maximum value of 2.0 to alleviate gradient explosion. 
The training batch size is 32 and the epoch number is 15.
We set the minimum span $k$ of connections to 2.
Moreover, we set the iteration number of pseudo-labeled data selection to 1.
The confidence threshold $\epsilon$ is 0.98.

\subsection{Results}
The approaches in this work can be thought of as a permutation of data, parsers, and filtering methods.
There are two sets of data, the syntactic dependency treebank $\mathcal{S}$ ($\mathcal{S}_\tau$), and the large-scale unlabeled dialogue data $\mathcal{D}$.
Also, there are two types of transformations $\textit{PreTran}$ and $\textit{PostTran}$.\footnote{Predictions for all methods can be processed by $\textit{PostTran}$. For consistency and convenience, our reported results are with $\textit{PostTran}$. Appendix \ref{sec:post} shows more details.}
For brevity, we simplify the $\textrm{Parser-S}$ to $\textit{f}_s$ and the $\textrm{Parser-T}$ to $\textit{f}_t$.
The pseudo-labeled data selection is denoted by a function $\eta$.
We record and analyze the experimental results of zero-shot and few-shot settings.

\noindent
\textbf{Zero-shot.}
\begin{table}[t]
\centering
\begin{tabular}{lccc}
\toprule
\textbf{Training Data} & \textbf{Inner} & \textbf{Inter} \\
\midrule
$\mathcal{S}$                      & 83.14          & 49.94                               \\
$\mathcal{S}_\tau$                 & 83.16 & 49.71                 \\
\hdashline
$\textit{f}_s\left(\mathcal{D}\right)$                           & 82.13          & 48.76                              \\
$\textit{f}_t\left(\mathcal{D}\right)$                            & 82.14          & 48.38                           \\
$\left( \textit{f}_s + \textit{f}_t \right) \left(\mathcal{D}\right)$                       & 82.46          & 49.03                        \\
\hdashline
$\mathcal{S} \hspace{0.4em} + \textit{f}_s\left(\mathcal{D}\right)$                            & 82.48          & 48.84      \\
$\mathcal{S}_\tau+ \textit{f}_t\left(\mathcal{D}\right)$                           & 82.68          & 49.02           \\
$\mathcal{S} \hspace{0.4em} + \eta \left( \textit{f}_s \left(\mathcal{D}\right) \right)$                     & 84.14 & 50.47         \\
$\mathcal{S}_\tau+\eta \left( \textit{f}_t \left(\mathcal{D}\right) \right)$                       & 84.13          & 50.27           \\
$\mathcal{S} \hspace{0.4em} + \eta \left( \textit{f}_s \left(\mathcal{D}\right) + \textit{f}_t \left(\mathcal{D}\right) \right)$           & 84.24          & 50.62      \\
$\mathcal{S}_\tau+ \eta \left( \textit{f}_s \left(\mathcal{D}\right) + \textit{f}_t \left(\mathcal{D}\right) \right)$           & \textbf{84.34}          & \textbf{50.78}      \\
\bottomrule
\end{tabular}
\caption{The test results under the zero-shot setting. We divide the results by a dashed line according to the source of training data. ``Inner'' represents the inner-EDU dependency, and ``Inter'' is the inter-EDU one.}
\label{tab:zero}
\end{table} 
Since the trends in the UAS and LAS are roughly consistent, here we only report the LAS. 
Details are recorded in Appendix \ref{sec:post}.
As shown in Table \ref{tab:zero}, the parser trained on syntactic treebank $\mathcal{S}$ and transformed treebank $\mathcal{S}_\tau$ achieve similar performances in inner-EDU parsing.
This shows how little our dependency transforming method disrupts the original syntactic structure.
It is intuitive that the parser independently trained on pseudo-labeled dialogue data $\mathcal{D}$ performs not well.
The parsers trained on the mergers of $\mathcal{S}$ and $\eta \left( \mathcal{D} \right)$ obtain the highest syntax parsing scores, demonstrating the usefulness of our data selection.

For the inter-EDU dependency parsing, it can be observed that the transformed treebank $\mathcal{S}_\tau$ makes the parser perform better.
Similar to inner-EDU parsing, the direct use of $\mathcal{D}$ performs poorly.
Although the ensemble of $\textrm{Parser-S}$ and $\textrm{Parser-T}$ can bring some performance gains, they are limited.
The combination of $\mathcal{S}$ ($\mathcal{S}_\tau$) and $\eta \left( \mathcal{D} \right)$ is useful.
The parser trained on $\mathcal{S}_\tau+\eta \left( \textit{f}_s \left(\mathcal{D}\right) + \textit{f}_t \left(\mathcal{D}\right) \right)$ achieves best inner-utterance performance.
It demonstrates the effectiveness of our signal-based dependency transformation and multi-view data selection.


\noindent
\textbf{Few-shot.}
We conduct few-shot (5, 10, 20, 50) experiments, incorporating the few labeled samples with those typologies of training data in the above zero-shot setting.
For clarity, we also present only the LAS.
Appendix \ref{sec:detail-few} gives more details.

As can be observed in Table \ref{tab:few}, our approaches significantly outperform parsers trained on only a small number of labeled training samples.
The performance improvement of inter-EDU parsing is more pronounced than the one of inner-EDU parsing, probably thanks to the fact that the former is a more difficult task with larger room for enhancement.
It can also be observed that as the annotated training data increases, the performance of inner-EDU parsing gradually reaches a higher level, and the improvement achieved by introducing augmented data becomes confined.
Nevertheless, parsers trained on the data with augmentation $\mathcal{S}_\tau+\eta \left( \textit{f}_s \left(\mathcal{D}\right) + \textit{f}_t \left(\mathcal{D}\right) \right)$ achieves highest scores in the majority of few-shot settings, showing the effectiveness of the proposed method.
\begin{table*}[t]
\centering
\begin{tabular}{l|cc|cc|cc|cc}
\hline
\multirow{2}{*}{\textbf{Augmented data}} & \multicolumn{2}{c|}{\textbf{5}}  & \multicolumn{2}{c|}{\textbf{10}} & \multicolumn{2}{c|}{\textbf{20}} & \multicolumn{2}{c}{\textbf{50}} \\
\cline{2-9}
                                & \textbf{Inner} & \textbf{Inter} & \textbf{Inner} & \textbf{Inter} & \textbf{Inner} & \textbf{Inter} & \textbf{Inner} & \textbf{Inter} \\ 
                                \hline
Null                            & 40.77   & 25.61  & 55.83  & 30.98   & 70.44  & 42.30  & 82.64   & 51.00      \\
$\mathcal{S}$                      & 85.01   & 50.02  & 85.83  & 50.98   & 86.80  & 52.04  & 88.20   & 54.42      \\
$\mathcal{S}_\tau$                      & 85.05   & 50.09  & 85.69  & 51.13   & 86.77  & 52.01  & \textbf{88.28}   & 54.53     \\
$\mathcal{S} \hspace{0.4em} + \eta \left( \textit{f}_s \left(\mathcal{D}\right) \right)$                       & 85.43   & 50.58  & 85.70  & 51.56   & \textbf{87.03}  & 52.51  & 88.18   & 55.50      \\
$\mathcal{S}_\tau+\eta \left( \textit{f}_t \left(\mathcal{D}\right) \right)$                      & 85.22   & 50.78  & 85.81  & 51.61   & 86.75   & 52.69  & 88.12  & 55.61       \\
$\mathcal{S}_\tau+\eta \left( \textit{f}_s \left(\mathcal{D}\right) + \textit{f}_t \left(\mathcal{D}\right) \right)$                   & \textbf{85.55}   & \textbf{51.12}  & \textbf{85.88}  & \textbf{51.88}   & 86.92  & \textbf{52.86}  & 88.20   & \textbf{55.73}   \\ 
\hline
\end{tabular}
\caption{The test results under the few-shot settings. Here we report the LAS.}
\label{tab:few}
\end{table*}
\section{Discussion}

\noindent
\textbf{What are the advantages of introducing discourse dependency compared to only syntax?}
Figure \ref{fig:example} has illustrated the superiority of introducing discourse dependency in two aspects.
Relationships between utterances can be expressed by discourse dependency.
Also, it represents a hierarchical structure and enables a more nuanced connection between EDUs.

\begin{figure}[t]
    \centering
    \includegraphics[width=0.49\textwidth]{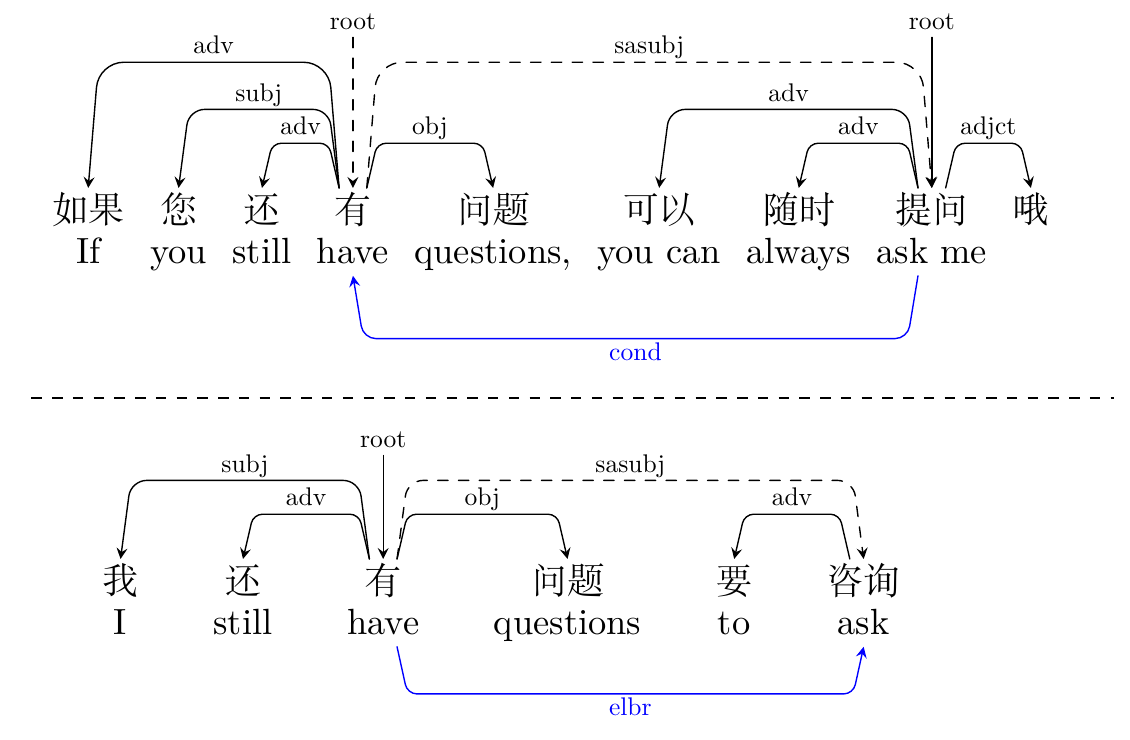}
    \caption{Two examples, where the above has a different rhetorical structure than the below, while they are almost isomorphic based on syntactic dependencies.}
    \label{fig:comp}
\end{figure}

Figure \ref{fig:comp} gives another sample, the syntactic dependency suffers from the problem of somorphism.
Those two sentences contain different semantic information, while their two syntactic subtrees are simply connected by ``sasubj'', resulting in their highly analogous architecture.
According to RST dependency, the two subtrees of the first sentence are linked by ``cond'' (condition) from right to left, and the two below are connected by ``elbr'' (elaboration) from left to right.
The inclusion of discourse dependency can alleviate the dilemma of somorphism by expressing high-level hierarchies.

\noindent
\textbf{How much overlap between partial syntactic dependencies and inter-EDU ones?}
Figure \ref{fig:example} provides an example that demonstrates the potential of transforming syntactic dependencies into inter-EDU ones.
Here we present a quantitative analysis.

\begin{figure*}[t]
\centering
\begin{subfigure}{0.3\textwidth}
    \includegraphics[width=\textwidth]{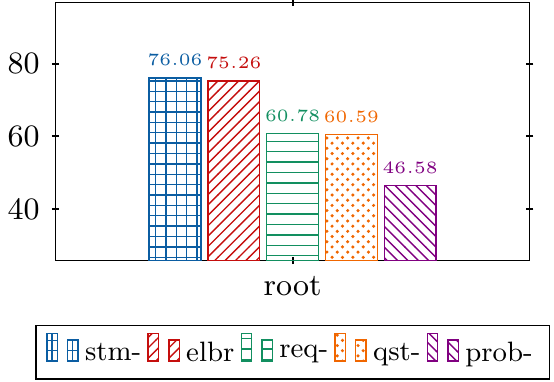}
    \label{fig:first}
\end{subfigure}
\hfill
\begin{subfigure}{0.3\textwidth}
    \includegraphics[width=\textwidth]{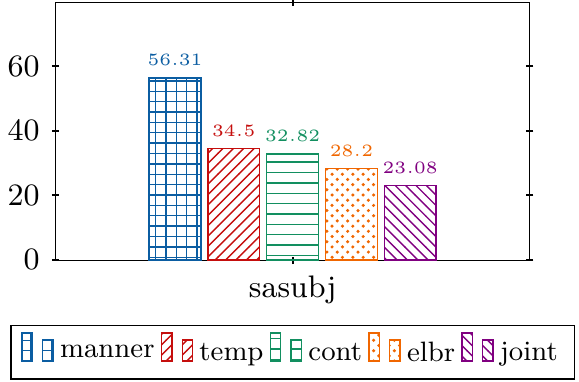}
    \label{fig:second}
\end{subfigure}
\hfill
\begin{subfigure}{0.3\textwidth}
    \includegraphics[width=\textwidth]{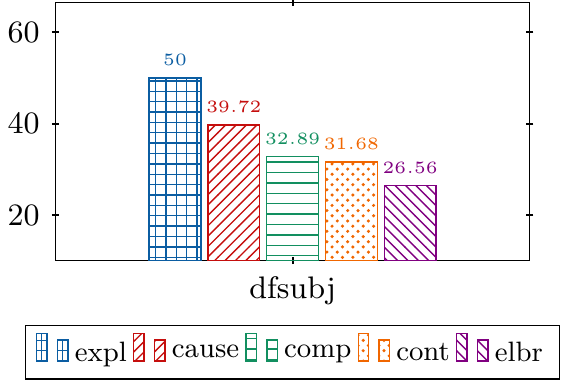}
    \label{fig:third}
\end{subfigure}
\caption{Top-5 matching labels and scores of ``root'', ``sasubj'', and ``dfsubj''.}
\label{fig:matching}
\end{figure*}

We present a matching score to measure the extent of overlap.
We first obtain syntactic predictions by $\textrm{Parser-S}$.
Then we replace those dependencies that bear a specific syntactic label with an inter-EDU class. 
Next, we compute the LAS for that class, as the matching score between the syntactic and inter-EDU labels.
Figure \ref{fig:matching} illustrates the top-5 matching scores of syntactic labels ``root'', ``sasubj'', and ``dfsubj''.
It can be seen that these syntactic labels have a respectable consistency with certain inter-EDU classes, revealing the potential for dependency conversion.
Interestingly, different syntactic labels are matched with different inter-EDU classes, indicating the overlap of specific syntactic structures with certain discourse structures.

\begin{figure}[t]
    \centering
    \includegraphics[width=0.48\textwidth]{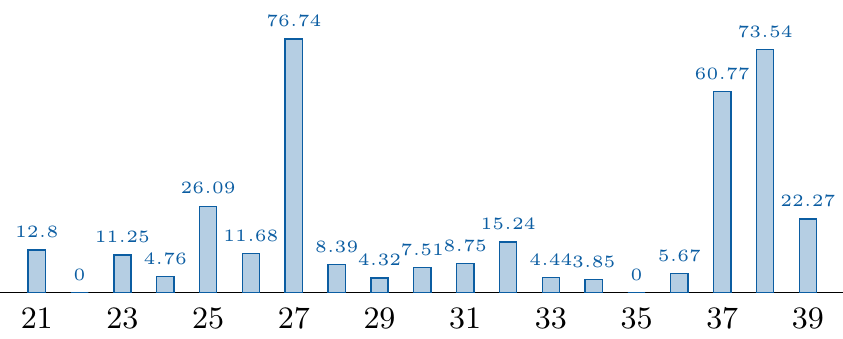}
    \caption{Matching score of signals. The numbers below sentences indicate the signals detected, in the order given in Appendix \ref{sec:dep}.}
    \label{fig:signal_match}
\end{figure}

\noindent
\textbf{How broadly can inter-EDU signals be reflected by pre-defined signal words?}
As mentioned above, certain words can reflect the semantic role of EDU.
We calculate the consistency of signal words and the relationship labels on EDUs to quantify the extent.
Given an EDU, we directly assign its label by the signal corresponding to the pre-defined signal word it contains.
Then, we compute the accuracy of each inter-EDU label as the matching score of the relevant signals.
It can be observed in Figure \ref{fig:signal_match} that some labels such as ``elbr'' (27), ``qst-ans'' (37), and ``stm-rsp'' (38) can be strongly reflected by the pre-defined signal words.
Some labels that do not contain significant signals in EDUs are maintained at low scores, especially ``bckg'' (22) and ``topic-chg'' (35), indicating that there is much room for improvement.

\begin{figure}[t]
    \centering
    \includegraphics[width=0.48\textwidth]{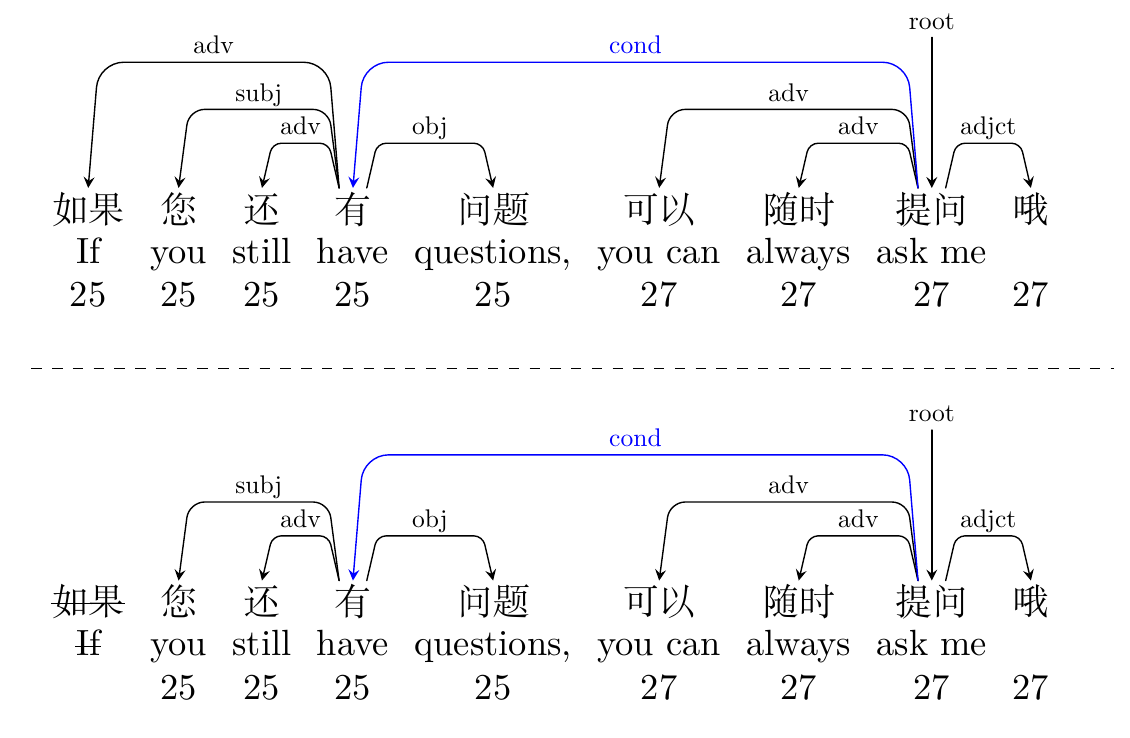}
    \caption{Two cases, where the above is with the explicit signal word ``If'' and the below is without it. The numbers represent the labels in the corresponding order.}
    \label{fig:signal}
\end{figure}

\noindent
\textbf{Is the MLM-based signal detection method able to detect implicit signal?}
Our inter-EDU signal detection method applies an MLM paradigm, which tries to recover the dropped signal word during the training stage.
Intuitively, this method should have the capacity to predict those implicit signals.
Figure \ref{fig:signal} gives two examples to prove that.
It can be seen that even when ``if'' is removed, the approach of MLM-based signal detection can still predict the ``cond'' signal, which is denoted as a 25 number.
Based on this, the parser can correctly predict the ``cond'' relations between two EDUs.

\begin{figure}[t]
    \centering
    \includegraphics[width=0.46\textwidth]{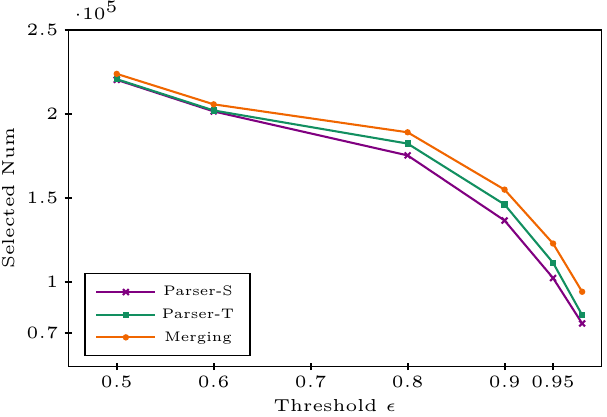}
    \caption{Quantity variation of filtered instances.}
    \label{fig:threshold}
\end{figure}

\noindent
\textbf{How does the reserved size of pseudo-labeled data change by confidence thresholds?}
We vary the threshold $\epsilon$ from 0.5 to 0.98 to investigate the changes in the amount of data selected.
Figure \ref{fig:threshold} illustrates the trend.
Interestingly, the size of selected data is still large even though the threshold is greater than 0.9, illustrating the high confidence in dependency parsing.
It can also be observed that the increase in merged data becomes more pronounced as the threshold increases.
Intuitively, as the amount of data selected decreases, fewer data will overlap.

\noindent
\textbf{Why not leverage Chinese RST treebanks to train a parser?}
To date, there exist two Chinese RST treebanks, the RST Spanish-Chinese Treebank \citep{cao-etal-2018-rst} and GCDT \citep{peng-etal-2022-gcdt}.
It is natural to use them to provide supervised signals for inter-EDU parsing.
In our prior experiments, we find that directly using these treebanks for training a parser leads to unsatisfactory performance.
One possible reason is that their annotation style is not aligned with ours. 
For instance, GCDT splits sentences very finely and uses lots of ``same-unit'' relations to align to the English style, fitting their multilingual tasks.
This occurs not in our data as it tends to cause an incomplete structure within the EDU.
Furthermore, GCDT lacks some of the label categories in our data, especially some common relations such as ``statement-response''.
In addition, inconsistencies in data distribution can also contribute to underperformance.
\section{Related Work}
\noindent
\textbf{Dependency parsing.}
To date, there are several Chinese dependency paradigms and the corresponding treebanks \citep{xue-2005-penn, che-2012-chinese, mcdonald-etal-2013-universal, qiu-2014-multi}.
These works mostly focus on sentence-level dependency parsing, while the document-level one is conspicuous by its paucity.
\citet{li-etal-2014-text} adopt a dependency parsing paradigm for discourse parsing, while its EDU-wise pattern neglects to parse inside EDUs.
We propose a unified schema, which includes both word-wise dependencies within and between EDUs that organize whole dialogues into tree structures.

\noindent
\textbf{Dialogue parsing.}
Discourse structures can be expressed by several theories, e.g., RST \citep{mann-1987-rhetorical, mann-1988-rhetorical}, SDRT \citep{asher-2003-logics}, and PTDB \citep{prasad-etal-2008-penn}.
The investigation of dialogue-level discourse parsing is still in its early stage.
\citet{afantenos-etal-2015-discourse, asher-etal-2016-discourse} build an annotated corpus STAC based on SDRT \citep{asher-2003-logics}, using a dependency paradigm for annotation of relationships between EDUs.
\citet{li-etal-2020-molweni} present the Molweni dataset, contributing discourse dependency annotations.
These two data are annotated at the granularity of EDUs and, to accommodate multi-party dialogue scenarios, their annotations are based on SDRT that fits the graph structure.
Differently, in line with common syntactic dependencies, the inter-EDU part of our dialogue-level dependency is word-wise and references RST \citep{carlson-etal-2001-building}, which organizes a text into the tree structure.
Furthermore, the prevalent corpus for dialogue dependency parsing is in English, and our dataset fills a lacuna in the Chinese corpus.

\noindent
\textbf{Weakly supervised learning.}
It is challenging to predict new classes that are unobserved beforehand \citep{norouzi-2014-zero}.
PLMs \citep{devlin-etal-2019-bert, clark-etal-2020-electra, brown-2020-language} can address this challenge through language modeling combined with label mapping, and prompt-based learning can stimulate this ability \citep{liu-etal-2021-prompt, ouyang-2022-training, lang-2022-co}.
Inspired by that, we adopt an MLM-based method to sense inter-EDU signals and map them to unseen dependencies.
Furthermore, our approach employs single-view and multi-view data selection, borrowing from self-training \citep{scudder-1965-probability} and co-training \citep{blum-1998-combining}, which are used for growing a confidently-labeled training set.
\section{Conclusion}
We presented the first study of Chinese dialogue-level dependency parsing.
First, we built a high-quality treebank named CDDT, adapting syntactic \citep{jiang-etal-2018-supervised} and RST \citep{carlson-etal-2001-building} dependencies into inner-EDU and inter-EDU ones.
Then, we conducted manual annotation and reached 850 labeled dialogues, 50 for training and 800 for testing.
To study low-resource regimes, we leverage a syntactic treebank to get inner-EDU dependencies and induce inter-EDU ones.
We employed an MLM method to detect the inter-EDU signals of each EDU and then assign the detected signals to the arcs between EDUs.
Furthermore, we exploited single-view and multi-view approaches for pseudo-labeled sample filtering.
Empirical results suggest that our signal-based method can achieve respectable performance in zero-shot and few-shot settings, and pseudo-labeled data utilization can provide further improvement.

\section*{Limitations}
This study suffers from four limitations.
The first limitation is that even though we annotated 850 dialogues manually, which includes almost 200,000 dependencies, there is still room for improvement in the total number of labeled dialogues.
The second one is that our parsing method of inter-EDU in the inter-utterance situation is simplistic and straightforward, and it can not cover certain difficult labels.
It is desirable to propose a more elegant and comprehensive approach.
The third is somewhat analogous to the second.
In the future, we should propose an end-to-end method that replaces the current approach, which consists of several processing steps.
The last one is about our pseudo-labeled data selection method.
It could be interesting to investigate the iterative process.

\section*{Ethics Statement}
We build a dialogue-level dependency parsing corpus by crowd annotations.
The raw dialogue data is obtained from an open source.
Besides, we remove information relating to user privacy.
The annotation platform is developed independently by us.  
All annotators were properly paid by their efforts. 
This dataset can be employed for dialogue-level dependency parsing in both zero-shot and few-shot setting as well as in any other data settings.

\section*{Acknowledgement}
We would like to thank the anonymous reviewers for their constructive comments, which help to improve this work.
This research is supported by the National Natural Science Foundation of China (No. 62176180).

\bibliography{anthology,custom}
\bibliographystyle{acl_natbib}

\appendix
\section{Brief Introduction of RST}
\label{sec:rst}
As a prevalent theory in discourse parsing, RST has been studied since early \citep{mann-1987-rhetorical, mann-1988-rhetorical} and has been broadly developed \citep{carlson-etal-2001-building, carlson-2001-discourse, soricut-marcu-2003-sentence}.
Within the RST framework, a clause is primarily regarded as an EDU, and its boundaries are determined using lexical and syntactic indicators. 
Each EDU involved in a relationship is attributed with a rhetorical status or nuclearity assignment, which characterizes its semantic role in the overall discourse structure.

The RST framework distinguishes between two types of relations: mononuclear and multinuclear.
Mononuclear relations consist of two units, namely the nucleus and the satellite, whereas multinuclear relations involve two or more units of equal importance.
A total of 53 mononuclear and 25 multinuclear relations can be used for tagging the RST corpus. 
These 78 relations can be categorized into 16 distinct classes that share common rhetorical meanings. 
Furthermore, to establish a clear structure within the tree, three additional relations are utilized, namely textual-organization, span, and same-unit.
Our inter-EDU dependency follows most of the coarse-grained relations, and the ``topic-comment'' relation is refined in four classes to fit the dialogue scenario.

\section{Dependencies for Dialogue}
\label{sec:dep}
Different from \citet{carlson-etal-2001-building}, we weaken the concept of nucleus and satellite in our dialogue-level dependencies. 
For mononuclear relations, an arc is emitted from the nucleus to the satellite. 
For the multinuclear ones, the arc is always emitted from above (left) to below (right).
The landing point of the arc is the core semantic word (always a predicate) of a discourse unit.
To satisfy the single root and single head constraints, we keep only the dummy root of the first utterance, and the ones of the other utterances are replaced by inter-utterance links.
To adapt the customer service scenario in our dialogue data, we add the ``req-proc'' label corresponding to a situation, where one party proposes a requirement, and the other proposes to handle it.
A model is encouraged to identify the more difficult relationships.
Thus, when annotators find that two relationships appear to be appropriate, the more difficult one should be chosen.

Table \ref{tab:details} shows the meaning and quantity of each label.
It can be seen that the labels of ``root'', ``obj'', ``att'', ``adv'', ``adjct'', and ``punc'' are with leading numbers in syntax dependencies.
In discourse dependencies, the ``elbr'' label is the most numerous of the inner-utterance dependencies, and the ``stm-rsp'' is the most in quantity in inter-utterance ones.

\begin{table}[t]
\centering
\small
\begin{tabular}{llrr}
\hline
\textbf{Label} &  \textbf{Meaning} & \textbf{Train} & \textbf{Test}\\
\hline
root & root & 1167 & 20086 \\
sasubj-obj & same subject, object & 14 & 325 \\
sasubj & same subject & 119  & 2234 \\
dfsubj & different subject & 21 & 373 \\
subj & subject & 693 & 11365 \\
subj-in & inner subject & 29 & 425 \\
obj & object & 1264 & 22752 \\
pred & predicate & 84 & 1508 \\
att & attribute modifier & 976 & 18685 \\
adv & adverbial modifier & 1472 & 25073 \\
cmp & complement modifier & 196 & 3307 \\
coo & coordination & 85 & 1403 \\
pobj & preposition object & 198 & 3540 \\
iobj & indirect-object & 15 & 433 \\
de & de-construction & 174 & 2714 \\
adjct & adjunct & 1158 & 20799 \\
app & appellation & 98 & 1569 \\
exp & explanation & 79 & 1354 \\
punc & punctuation & 1238 & 20966 \\
frag & fragment & 10 & 127 \\
repet & repetition & 39 & 534 \\
\hdashline
attr & attribution & 26 & 336 \\
bckg & background & 33 & 741 \\
cause & cause & 27 & 391 \\
comp & comparison & 8 & 210 \\
cond & condition & 27 & 755 \\
cont & contrast & 14 & 368 \\
elbr & elaboration & 696 & 12199 \\
enbm & enablement & 14 & 286 \\
eval & evaluation & 4 & 139 \\
expl & explanation & 18 & 333 \\
joint & joint & 26 & 480 \\
manner & manner-means & 7 & 105 \\
rstm & restatement & 25 & 473 \\
temp & temporal & 41 & 597 \\
tp-chg & topic-change & 6 & 74 \\
prob-sol & problem-solution & 26 & 441 \\
qst-ans & question-answer & 229 & 3701 \\
stm-rsp & statement-response & 381 & 6617 \\
req-proc & requirement-process & 63 & 1127 \\
\hline
\end{tabular}
\caption{Statistics of all the dependency relations.}
\label{tab:details}
\end{table}

\section{Implemented Details of MLM}
\label{sec:imp_mlm}
We use the large-scale unlabeled dialogue data $\mathcal{D}$ to train a signal detection model.
The model contains an encoder layer of ELECTRA \citep{clark-etal-2020-electra} and a linear layer as the decoder to project the hidden vector to a vocabulary space.
Inspired by prompt-based learning, we set a template as ``The word that expresses the signal of discourse dependency is: [mask] [mask] [mask]'' in Chinese.
The template is as a prefix of input $\boldsymbol{x}$ to obtain the prompt $\boldsymbol{x}'$.
As there are many signal words and they are in Chinese, it is difficult to show them and thus we present them in publicly available code.
We average the output probabilities in [mask] positions to obtain the signal word distribution.
The probability of dropping a word is 0.2, and the one of dropping the signal word is 0.7.
We set the epoch number to 2.
The other hyper-parameters are the same as section \ref{sec:setting}.

\section{Impact of Post-Transformation}
\label{sec:post}
Table \ref{tab:post} shows the impact of $\textit{PostTran}$.
It can be observed that $\textit{PostTran}$ can benefit the most situations of discourse dependency parsing, both without and with $\textit{PreTran}$.
We think this is because $\textit{PostTran}$ determines some ambiguous predictions by detected signals.
Meanwhile, we find that $\textit{PostTran}$ sometimes impairs the performance of inner-EDU parsing.
This may be owing to incorrect signals or insufficiently comprehensive rules.

\begin{table*}[t]
\centering
\begin{tabular}{lccccc}
\toprule
\multirow{2}{*}{\textbf{Training Data}} & \multirow{2}{*}{\textbf{Method}} & \multicolumn{2}{c}{\textbf{Inner}}      & \multicolumn{2}{c}{\textbf{Inter}} \\ \cmidrule(r){3-4}  \cmidrule(r){5-6}
                               &                         & UAS            & LAS            & UAS               & LAS               \\ \midrule
\multirow{2}{*}{$\mathcal{S}$}    & -                       & 87.19          & 83.77          & /                 & /                 \\
                               & +\textit{PostTran}               & 87.37 & 83.14          & \textbf{65.73}             & \textbf{49.94}             \\
\multirow{2}{*}{$\mathcal{S}_\tau$}    & -                       & \textbf{87.38}          & \textbf{83.95} & 65.57             & 49.78             \\
                               & +\textit{PostTran}               & 87.32          & 83.16          & 65.48    &  49.71    \\ \hdashline
\multirow{2}{*}{$\textit{f}_s\left(\mathcal{D}\right)$}          & -                       & 86.13          & 82.90          & 64.21             & 48.50             \\
                               & +\textit{PostTran}               & 86.18          & 82.13          & 64.31             & 48.76             \\
\multirow{2}{*}{$\textit{f}_t\left(\mathcal{D}\right)$}          & -                       & 86.11          & 82.83          & 63.84             & 48.23             \\
                               & +\textit{PostTran}               & 86.11          & 82.14          & 63.93             & 48.38             \\
\multirow{2}{*}{$\left( \textit{f}_s + \textit{f}_t \right) \left(\mathcal{D}\right)$}      & -                       & 86.48          & 82.46           & 64.58             & 48.92            \\
                               & +\textit{PostTran}               & 86.48          & 82.46          & 64.75             & 49.03             \\ \hdashline
\multirow{2}{*}{$\mathcal{S} \hspace{0.4em} + \textit{f}_s \left ( \mathcal{D} \right)$}         & -                       & 86.20          & 82.99          & /                 & /                 \\
                               & +\textit{PostTran}               & 86.48          & 82.48          & 64.76             & 48.84             \\
\multirow{2}{*}{$\mathcal{S}_\tau + \textit{f}_t \left ( \mathcal{D} \right)$}         & -                       & 86.73          & 82.85          & 64.88             & 48.62             \\
                               & +\textit{PostTran}               & 86.72          & 82.68          & 64.93             & 49.02             \\
\multirow{2}{*}{$\mathcal{S} \hspace{0.4em} + \eta \left( \textit{f}_s \left ( \mathcal{D} \right) \right)$}      & -                       & 87.97          & 84.81          & /                 & /                 \\
                               & +\textit{PostTran}               &  88.05 & 84.14          & 66.29             & 50.47             \\
\multirow{2}{*}{$\mathcal{S}_\tau + \eta \left( \textit{f}_t \left ( \mathcal{D} \right) \right)$}      & -                       & 88.13          & 84.92 & 65.83            & 50.20            \\
                               & +\textit{PostTran}               & 88.08          & 84.13          & 65.88             & 50.27             \\
\multirow{2}{*}{$\mathcal{S} \hspace{0.4em} + \eta \left( \textit{f}_s \left ( \mathcal{D} \right) + \textit{f}_t \left ( \mathcal{D} \right) \right)$} & -                       & 88.02          & 84.78          & 65.84             & 50.18             \\
                               & +\textit{PostTran}               & 88.12          & 84.24          & 66.34    & 50.62    \\
\multirow{2}{*}{$\mathcal{S}_\tau + \eta \left( \textit{f}_s \left ( \mathcal{D} \right) + \textit{f}_t \left ( \mathcal{D} \right) \right)$} & -                       & 88.09          & \textbf{84.98}          & 66.45             & 50.60             \\
                               & +\textit{PostTran}               & \textbf{88.22}          & 84.34          & \textbf{66.48}    & \textbf{50.78}    \\ 
\bottomrule
\end{tabular}
\caption{The impact of \textit{PostTran}.}
\label{tab:post}
\end{table*}


\section{Detailed Results of Few-shot Settings}
\label{sec:detail-few}
Table \ref{tab:detail-few} shows the details of results in few-shot settings.
The definitions of the symbols are consistent with those above.
It can be seen that the tends of UAS and LAS are extremely in line.

\begin{table*}[t]
\centering
\begin{tabular}{lcccc}
\hline
\multirow{2}{*}{\textbf{Training Data}} & \multicolumn{2}{c}{\textbf{Inner}}       & \multicolumn{2}{c}{\textbf{Inter}}                         \\ \cmidrule(r){2-3}  \cmidrule(r){4-5}
                               & UAS            & LAS            & \multicolumn{1}{c}{UAS} & \multicolumn{1}{c}{LAS} \\ \hline
5-shot                         & 44.51±4.48     & 40.77±4.84     & 35.41±3.76                   & 25.61±2.92                   \\
$+ \mathcal{S}$                          & 88.72          & 85.01          & 65.80                   & 50.02                   \\
$+ \mathcal{S}_\tau$                   & 88.76          & 85.05          & 65.88                   & 50.09                   \\
$+ \mathcal{S} \hspace{0.4em} + \eta \left( \textit{f}_s \left ( \mathcal{D} \right) \right)$              & 89.28          & 85.43          & 66.45                   & 50.58                   \\
$+ \mathcal{S}_\tau + \eta \left( \textit{f}_t \left ( \mathcal{D} \right) \right)$         & 88.98          & 85.22          & 66.56                   & 50.78                   \\
$+ \mathcal{S}_\tau + \eta \left( \textit{f}_s \left ( \mathcal{D} \right) + \textit{f}_t \left ( \mathcal{D} \right) \right)$   & \textbf{89.15} & \textbf{85.55} & \textbf{67.57}                   & \textbf{51.12}          \\ \hline
10-shot                        & 59.01±4.86     & 55.83±4.94     & 45.30±5.72                   & 30.98±3.86                   \\
$+ \mathcal{S}$                              & 89.34          & 85.83          & 66.94                   & 50.98                   \\
$+ \mathcal{S}_\tau$                     & 89.27          & 85.69          & 67.09                   & 51.13                   \\
$+ \mathcal{S} \hspace{0.4em} + \eta \left( \textit{f}_s \left ( \mathcal{D} \right) \right)$                 & 89.35          & 85.70          & 67.52                   & 51.56                   \\
$+ \mathcal{S}_\tau + \eta \left( \textit{f}_t \left ( \mathcal{D} \right) \right)$   & 89.45          & 85.81          & 67.47                   & 51.61                   \\
$+ \mathcal{S}_\tau + \eta \left( \textit{f}_s \left ( \mathcal{D} \right) + \textit{f}_t \left ( \mathcal{D} \right) \right)$   & \textbf{89.50} & \textbf{85.88} & \textbf{68.08}          & \textbf{51.88}          \\ \hline
20-shot                        & 73.51±1.31     & 70.44±1.28     & 55.51±2.70                   & 42.30±2.13                   \\
$+ \mathcal{S}$                              & 90.22          & 86.80          & 68.30                   & 52.04                   \\
$+ \mathcal{S}_\tau$                   & 90.19          & 86.77          & 68.21                   & 52.01                   \\
$+ \mathcal{S} \hspace{0.4em} + \eta \left( \textit{f}_s \left ( \mathcal{D} \right) \right)$               & \textbf{90.74} & \textbf{87.03} & 69.10                   & 52.51                   \\
$+ \mathcal{S}_\tau + \eta \left( \textit{f}_t \left ( \mathcal{D} \right) \right)$  & 90.16          & 86.75          & 68.97                   & 52.69                   \\
$+ \mathcal{S}_\tau + \eta \left( \textit{f}_s \left ( \mathcal{D} \right) + \textit{f}_t \left ( \mathcal{D} \right) \right)$   & 90.56          & 86.92          & \textbf{69.20}          & \textbf{52.86}          \\ \hline
50-shot                        & 85.36±0.22     & 82.64±0.25     & 66.39±0.98                   & 51.00±0.66                   \\
$+ \mathcal{S}$                             & 91.62          & 88.20          & 70.93                   & 54.42                   \\
$+ \mathcal{S}_\tau$                    & \textbf{91.75} & \textbf{88.28} & 71.00                   & 54.53                   \\
$+ \mathcal{S} \hspace{0.4em} + \eta \left( \textit{f}_s \left ( \mathcal{D} \right) \right)$                & 91.70          & 88.18          & 72.05                   & 55.50                   \\
$+ \mathcal{S}_\tau + \eta \left( \textit{f}_t \left ( \mathcal{D} \right) \right)$  & 91.68          & 88.12          & 72.35                   & 55.61                   \\
$+ \mathcal{S}_\tau + \eta \left( \textit{f}_s \left ( \mathcal{D} \right) + \textit{f}_t \left ( \mathcal{D} \right) \right)$   & 91.74          & 88.20          & \textbf{72.53}          & \textbf{55.73}          \\ \hline
\end{tabular}
\caption{The details of few-shot results.}
\label{tab:detail-few}
\end{table*}


\end{document}